\documentclass[letterpaper, 10 pt, journal]{IEEEtran}

\IEEEoverridecommandlockouts                              
\usepackage[rightcaption,raggedright]{sidecap}
\usepackage{wrapfig}
\usepackage{graphicx} 
\usepackage{epsfig} 
\usepackage{mathptmx} 
\usepackage{times} 
\usepackage{amsmath} 
\usepackage{amssymb}  
\usepackage{flushend}

\DeclareMathOperator*{\argmax}{arg\,max}

\title{A Framework for Estimating Long Term Driver Behavior}

\author{Vijay~Gadepally,
       and Ashok~Krishnamurthy
\thanks{V. Gadepally is with the Massachusetts Institute of Technology,
MA, USA. The work described in this article was performed at The Ohio State University, OH, USA. e-mail: vijayg [at] mit.edu.}
\thanks{A. Krishnamurthy is with the Renaissance Computing Institute, Chapel Hill, NC, 27517 USA, e-mail: ashok [at] renci.org.}
}

\begin{document}

\maketitle

\begin{abstract}

The authors present a cyber-physical systems study on the estimation of driver behavior in autonomous vehicles and vehicle safety systems. Extending upon previous work, the approach described is suitable for the long term estimation and tracking of autonomous vehicle behavior. The proposed system makes use of a previously defined Hybrid State System and Hidden Markov Model (HSS+HMM) system which has provided good results for driver behavior estimation. The HSS+HMM system utilizes the hybrid characteristics of decision-behavior coupling of many systems such as the driver and the vehicle, uses Kalman Filter estimates of observable parameters to track the instantaneous continuous state, and estimates the most likely driver state. The HSS+HMM system is encompassed in a HSS structure and inter-system connectivity is determined by using Signal Processing and Pattern Recognition techniques. The proposed method is suitable for scenarios that involve unknown decisions of other individuals, such as lane changes or intersection precedence/access. The long term driver behavior estimation system involves an extended HSS+HMM structure that is capable of including external information in the estimation process. Through the grafting and pruning of metastates, the HSS+HMM system can be dynamically updated to best represent driver choices given external information. Three application examples are also provided to elucidate the theoretical system.

\end{abstract}


\section{Introduction}
\label{intro}

One of the projects at the Center for Intelligent Transportation Research (CITR) at The Ohio State University is in addressing the Cyber Physical Systems (CPS) related problem of \emph{autonomous vehicles operating safely in mixed-traffic urban environments}. Cyber Physical Systems refer to the combination of a system's computational and physical elements. Examples of Cyber Physical Systems include, but are not limited to, robotic surgery, automated traffic control and autonomous vehicles. CPS has the ability to remove many of the expensive and dangerous hurdles required in common human tasks. One application area for CPS is in personal transportation through the development of autonomous or self-driving vehicles which have the potential to revolutionize transportation by removing humans from the driving loop. Autonomous vehicles have the ability to address some of the largest issues associated with personal and commercial transportation. Of course, as with any new technology, if there is to be widespread adoption of autonomous vehicles, it is important to look at the possible adoption path and determine challenges with this adoption path.

A likely adoption path for  autonomous vehicles will be through the gradual inclusion of either fully autonomous vehicles or vehicles with partial autonomous capabilities into the current transportation system. Such an environment, in which there are both autonomous and human-driven vehicles operating together, is known as a mixed-traffic urban environment (MUE). In such an environment, autonomous vehicles will interact with human-driven vehicles in their regular operation.  This interaction will require that autonomous vehicles have the ability to receive information about human driven vehicles - either through Vehicle-to-Vehicle (V2V) communication or using onboard sensors. With such information, the autonomous vehicle must be able to deduce the behavior of other vehicles. In particular, in this interaction between autonomous and human-driven vehicles, the task of successful estimation and prediction of the expected behavior of the other vehicles is essential to avoid and/or mitigate the threat of traffic accidents. Such requirements are also necessary in the development of vehicles with advanced safety systems. The ability of an autonomous vehicle, based on sensor or V2V input, determine the likely behavior of a human-driven vehicle is referred to as \textit{Driver Behavior Estimation}. 

A human-driven vehicle follows a continuous trajectory that is set by discrete decisions made by a human driver. For example, consider the driver decision ``Stop at Intersection.'' A driver may make this decision for a number of reasons, but given this decision, the driver's vehicle will follow an observable trajectory of reduction in velocity, deceleration, illumination of brake lights, and eventually coming to a stop. The estimation of vehicle/driver decision consists of observing the continuous vehicle variables and determining the driver state (or driver decision) that caused these observations. For the purpose of this article, the term ``driver'' is used to describe the combination of human operator and vehicle, and the behavior exhibited by this combination is referred to simply as \textit{driver behavior}. In order to accurately determine driver behavior, a framework capable of describing the qualitative and quantitative nature of driver behavior estimation is needed. Extending upon a previously designed system for driver behavior estimation is the main topic for this article.

In the previously developed framework, the driver decision-vehicle dynamics coupling is encapsulated in a Hybrid-State System (HSS) representation, which has been used in studies such as~\cite{KurtOzgunerITSC10, kurtjournal, gadepally2011driver}. This representation has been useful for a number of automotive-related research areas, such as autonomous vehicles, and is utilized for the analysis, modeling and estimation of the driver/vehicle behavior. The HSS provides an intuitive structure in which the continuous vehicle state and discrete driver state are represented as two layers that, combined, describe vehicle-driver coupling. The novelty of this structure, among other hybrid-state estimation techniques, is in the ability to exploit the uniformity of a vehicle model under various situations. Estimation and prediction of components of this system is done through borrowed tools from signal processing based state estimation. In the developed framework, the HSS provides the basis of the qualitative aspect of driver behavior estimation.

The quantitative aspect (HSS system interconnections) of driver behavior estimation is provided by a type of graphical models. The discrete state estimation that gives the driver decision state is carried out through pattern recognition and Hidden Markov Models (HMMs)~\cite{rabiner:1986z}. HMMs are doubly stochastic models that can identify the underlying relationship between observations and the hidden states that generate these observations. HMMs have had success in topics such as speech recognition~\cite{rabiner:1989m}, human behavior {\cite{Pentland1999a} and recognizing driving events~\cite{mitrovic2005reliable}. In our application, continuous vehicle observations are modeled as Gaussian Mixture Models for the learning and evaluation of HMMs. State estimation is done through pattern recognition techniques that relate continuous observations with likely driver states. Overall, the HSS provides the system architecture and HMMs define the quantitative relationship between system components. This combination of HSS and HMM is referred to as \textit{HSS+HMM}.

In~\cite{gadepally2014framework}, we presented results obtained for driver behavior estimation using the HSS+HMM system. The system presented in that article performs accurate driver behavior estimation around intersections. This system was expanded to include lane change maneuvers and highway events to better describe the long term behavior of a vehicle. Previous studies such as~\cite{miller1983long, varhelyi2004effects, wilde1976social} have only looked at aspects of long term driver bevavior. 

This article begins with an overview of the HSS+HMM system, in which we discuss the developed system along with the data collection experiment. Section~\ref{longterm} discusses the current system implementation along with limitations that hamper estimating long term driver behavior. Section~\ref{dynamicdss} discusses a proposed theoretical extension to the HSS+HMM system along with definitions. Finally, we discuss three example implementations of the proposed system and conclude the article.


\section{The HSS+HMM System}
\label{hsshmm}

The HSS+HMM system refers to the conjoining of two models - Hybrid State Systems and Hidden Markov Models. In order to estimate, track and predict the behavior of the vehicle and its driver, the interaction between the vehicle and the driver is captured in a Hybrid-State System (HSS) model, which consists of a discrete-state system (DSS) higher level and a continuous-state system (CSS) lower level, as seen in the left part of Figure~\ref{hss_hmm_rep}. An assumption of the system is that a driver reacts to discrete events, makes corresponding decisions on the higher level, and the vehicle follows continuous trajectories according to driver intention. This coupling of systems with different domain characteristics has been modeled as a HSS for a variety of applications, including hybrid-state controllers for autonomous vehicles~\cite{KurtOzgunerIFAC08}. The interaction between the modes or the states in the DSS is modeled as a Finite State Machine (FSM). For the application of interest, measurements made through V2V communication or on-board sensors provides an estimate of the CSS state.

Given a sequence of CSS state estimates, determination of the higher level DSS state is done by using Hidden Markov Models and pattern recognition techniques. Using a data collection process described in Section~\ref{datacollection}, numerous HMMs are trained using the Baum Welch method. Once models $\lambda_{1}, \lambda_{2}, ... , \lambda_{n}$, corresponding to \textit{n} different vehicle actions (such as a vehicle turning left, vehicle going straight, etc.) have been trained, we can perform state estimation using pattern recognition techniques. For a given observation sequence \textbf{O} = $\{$$o_{1}, o_{2}, ..., o_{t}$$\}$, the probabilities $P(O | \lambda_{i})$, $i=1, 2, ..., n$ are calculated using the forward algorithm. The highest likelihood probability corresponds to the estimated vehicle action at that time. Thus, to determine the state at some time, $t$:

\begin{eqnarray}\label{argmaxt} \mbox{$State(t) $} = \argmax_i P(o_{1}o_{2}...o_{t} | \lambda_{i}) \mbox{      $i = 1, ...., n$}  \end{eqnarray}

This process is summarized grapically in Figure~\ref{taskgraft_general}. The incoming observations correspond to estimates of the CSS state. The metastates (shaded in green) correspond to different vehicular events of interest that have been trained using actual data. The final compare step corresponds to equation~\ref{argmaxt} and the output estimated state is an estimate of the DSS state.

\begin{figure}[t]
\begin{center}
\includegraphics[width=8.9cm]{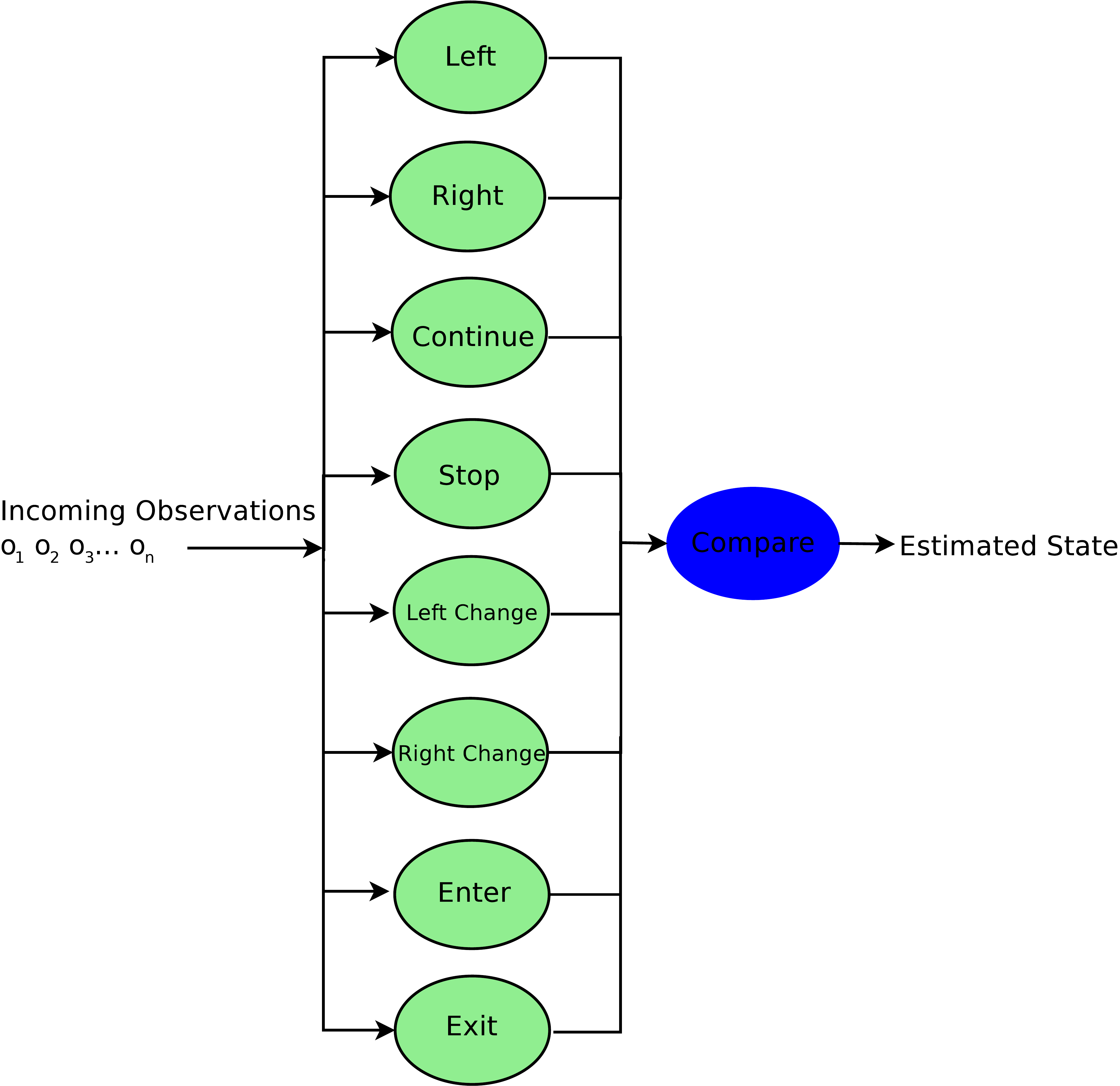}    
\caption{Graphical Representation of Possible Event Metastates and incoming CSS state estimates. Output estimated state corresponds to HSS+HMM state estimate.}  
\label{taskgraft_general}                                 
\end{center}                                 
\end{figure}

The DSS of the HSS contains states that represent driver intentions. The HMM based approach defines the relationship between \textit{easy to measure} continuous observations and a related metastate in the DSS. With such a framework, each state of the DSS, referred to as a \textit{metastate}, is in essence a HMM which corresponds to a vehicle event of interest. Figure~\ref{hss_hmm_rep} describes this relationship where metastate $S_{3}$ is shown to be a HMM with states \{$S^{1}_{3},S^{2}_{3}, S^{3}_{3}, S^{4}_{3}$\}.   These states correspond to the hidden states of the HMM (in this example, $N=4$). 

\begin{figure}[t]
\begin{center}
\includegraphics[width=7.1cm]{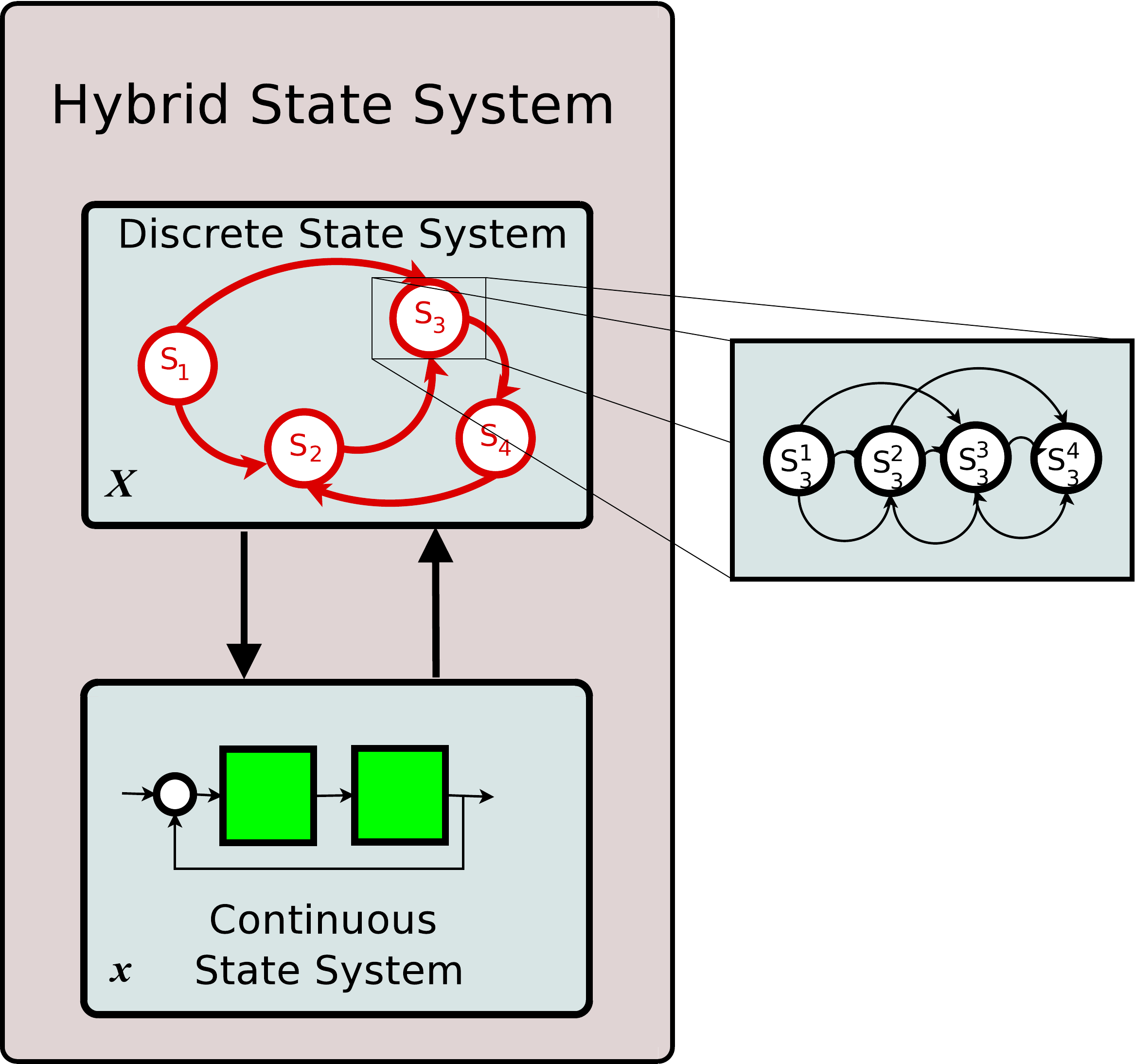}      
\caption{Zoomed in version of metastate $S_{3}$ in DSS. Each metastate is a Hidden Markov Model}  
\label{hss_hmm_rep}                                 
\end{center}                                 
\end{figure}

Within the HSS+HMM system, a change in the metastate of the DSS implies a change in each DSS state. Transitions between states of the DSS are determined by using equation~\ref{argmaxt} on the observation sequence. Consider a situation in which the DSS consists of four possible states: \{$S_{1}, S_{2}, S_{3}, S_{4}$\}. Suppose that each of these states is represented by a 4 hidden state HMM. Suppose now that a vehicle was observed to traverse these states of the DSS in the following order: \{$S_{1}, S_{2}, S_{3}, S_{4}$\} which is represented by the green lines in figure~\ref{states_traverse}. Since each state of the DSS is essentially represented by a HMM, and traversing through states of the DSS represents traversing through the metastates of the HMM, for illustration purposes, the red lines in figure~\ref{states_traverse} represents one of the possible transition sequences that represents the DSS states transition, namely: 

\begin{eqnarray} {S^{1}_{1},S^{3}_{1},S^{4}_{1},S^{1}_{2},S^{2}_{2},S^{3}_{2},S^{4}_{2},S^{1}_{3},S^{2}_{3},S^{4}_{3},S^{1}_{4},S^{2}_{4},S^{4}_{4}} \nonumber 
 \end{eqnarray}

\begin{figure}[t!]
\begin{center}
\includegraphics[width=8.4cm]{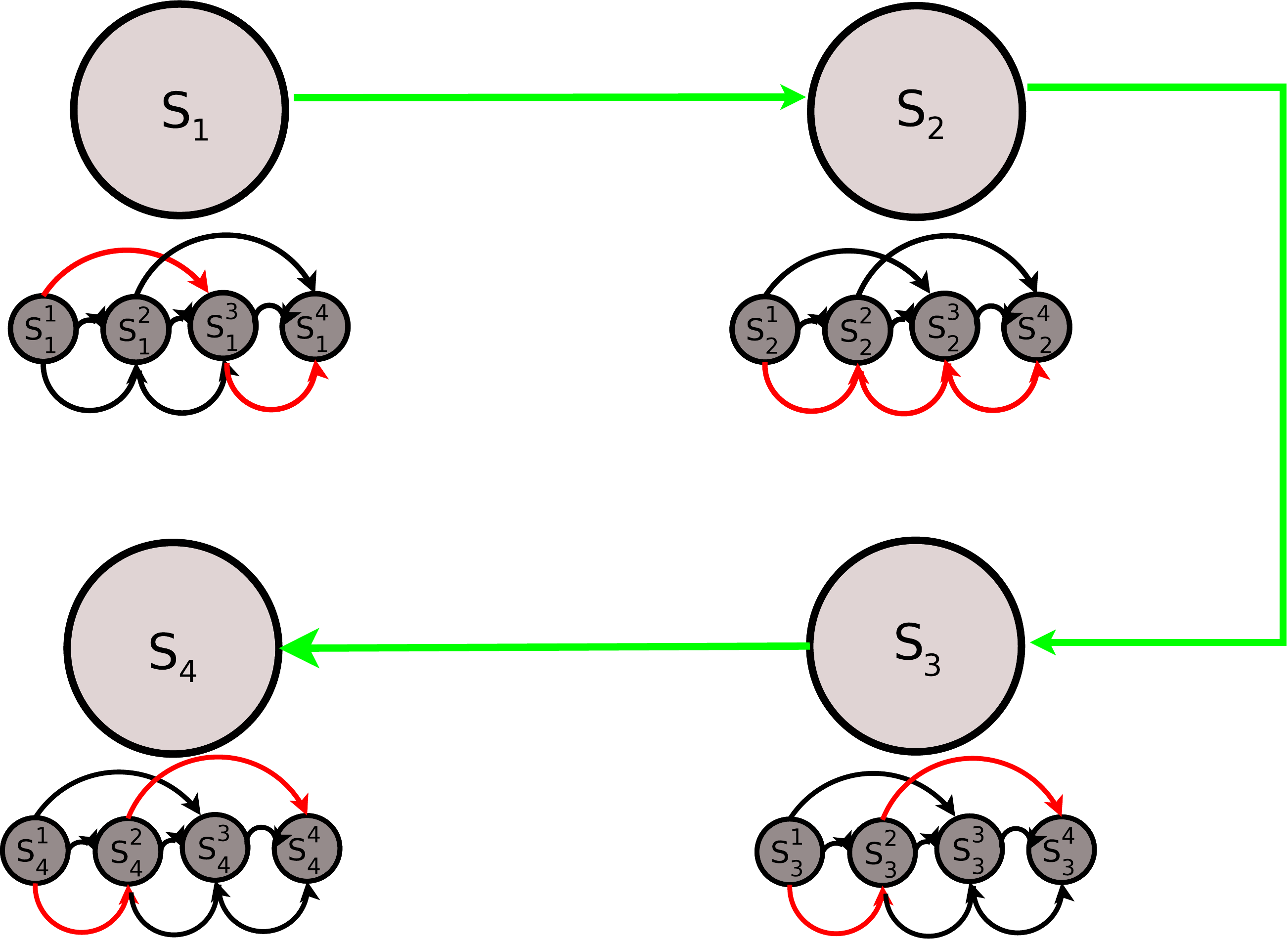}      
\caption{Possible traversal of DSS states. Green lines represent DSS metastate transitions, Red lines represent actual state transitions}  
\label{states_traverse}                                 
\end{center}                                 
\end{figure}

In order to accurately estimate and predict driver behavior, it is necessary to train a large set of DSS metastates that correspond to different vehicle events of interest. Additionally, after determining the relationship between DSS states (i.e., transition probabilities of the DSS FSM), one may use an algorithm such as the Viterbi Algorithm~\cite{Forney_Viterbi} to predict the most likely future state. A brief summary of the data collection procedure is given in the following section.

\subsection{Data Collection}
\label{datacollection}

In order to train the metastates of the HSS+HMM model, participants have been recruited to operate a sensor-fitted vehicle around the streets of Columbus, OH. Since our study concentrates on determining the relationship between easily observable continuous system estimates to estimate the state that describes the high level behavior of a driver we have designed the experiement to be representative of typical driving scenarios (regular traffic, dry roads, etc.). An implicit assumption for the experiment is that given the correct combination of sensors or V2V communication, one will be able to measure any of the CSS state estimates we are using in the study. For convenience and accuracy, we make use of the on-board instrumentation on a vehicle to measure these low-level continuous observations.

\begin{figure}[t]
\begin{center}
\includegraphics[width=8.4cm]{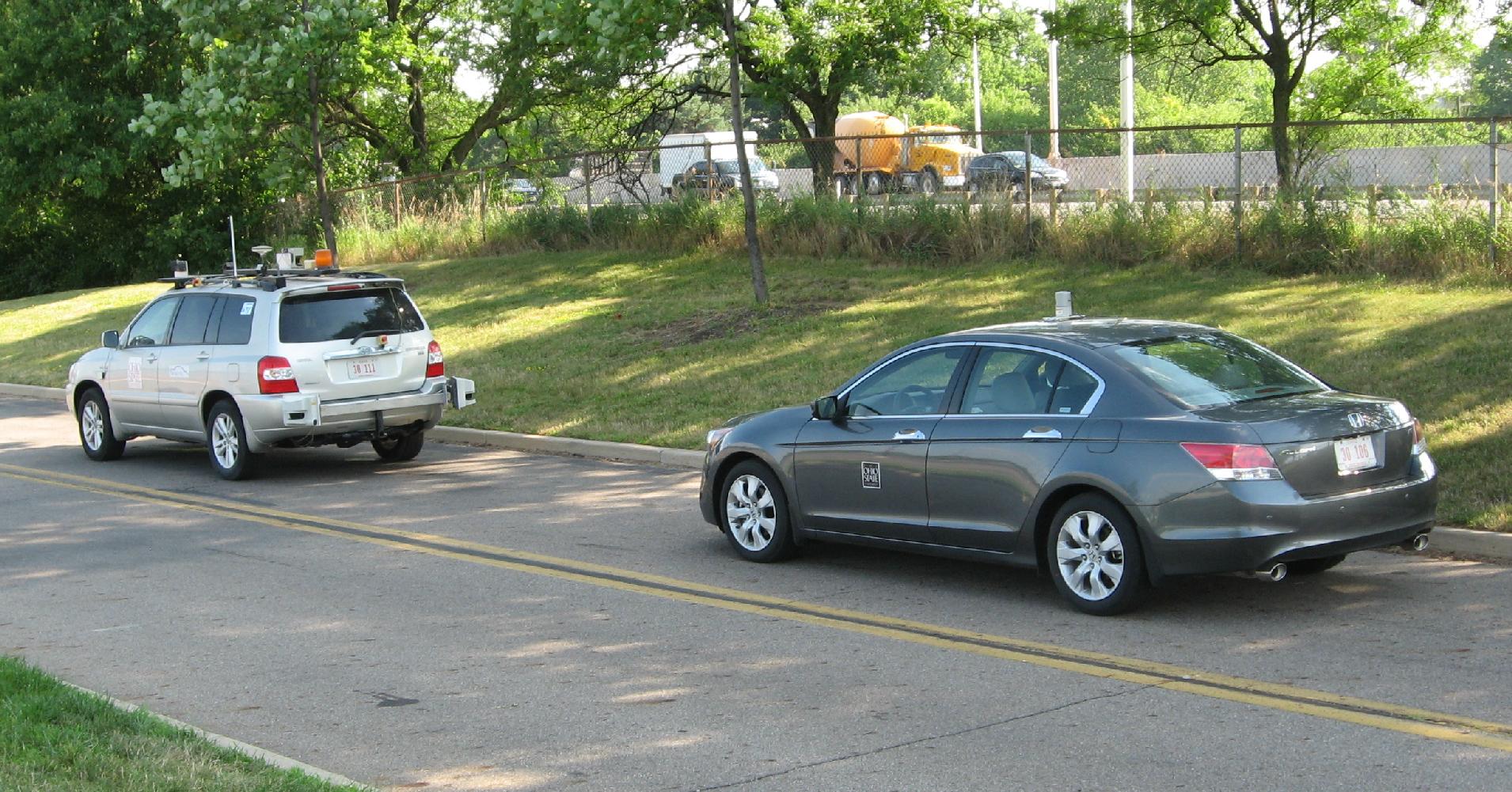}    
\caption{OSU CITR Vehicles in a convoy. The Honda Accord (Right) is used for Data Collection}  
\label{2012Honda}                                 
\end{center}                                 
\end{figure}

A 2012 Honda Accord (Figure ~\ref{2012Honda}) fitted with the following sensors has been used for data collection:

\begin{itemize}
\item \textbf{Novatel GPS Unit}. This sensor provides: GPS Latitude, GPS Longitude, Timestamp of reading, Altitude, Latitude and Longitude Estimated Standard Deviation, Horizontal Speed, Track over Ground (Yaw).
\item \textbf{Honda Accord CAN Bus}. This sensor provides: Timestamp of Reading, Yaw Rate, Lateral Acceleration, Throttle Pedal Position, Brake Pressure, Steering Wheel Angle, Speed, Engine RPM, Torque Converter RPM, Odometer, Headlights, Brake Lights, Throttle Peddle, Gear, Wiper, Turn Signals.
\item \textbf{Three HD cameras} that provide views of the front, left side, and right side of the vehicle.
\end{itemize}

Data collected from these sensors are fused into relevent feature vectors which are then used for model training. We concentrate on events of interest that occur near intersections, roads, and highways. For this purpose, we have limited the models to the following events:

\begin{enumerate}
\item Left Turn
\item Right Turn
\item Straight/Continue
\item Stop
\item Left Lane Change
\item Right Lane Change
\item Enter Highway
\item Exit Highway
\end{enumerate}

Additional experimental data was collected that validated the HSS+HMM system with respect to naturalistic driver behavior estimation. Details of the experiment and results is available in~\cite{gadepally2014framework}.


\section{Long Term Driver Behavior}
\label{longterm}

Consider the HSS+HMM metastate estimation stage for driver behavior estimation described in the previous section. Previous work on driver behavior estimation has concentrated on estimating the short term behavior (i.e, during a particular event or at a particular location). In many cases, there is a need to estimate or track a full vehicle driving sequence. In a full driving sequence, a driver goes through a series of events to accomplish a goal often through changing scenarios. For example, to drive to work from home, a typical sequence of driving events may be: back out of the garage, drive through city streets for a few miles, take a limited access highway for several miles, take an exit, drive through city streets, turn into a parking garage and finally park.

\subsection{Long Term Behavior Estimation Limitations}

HSS+HMM system provides very good driver behavior results when compared to other possible systems. One of the limitations of the HSS+HMM system is in describing the long term behavior of a vehicle. This limitation is largely due to two reasons: 1) Metastate likelihood probabilities reduce for long observation sequences and 2) The HSS+HMM system does not include external information into the estimation process.

\begin{figure}[h]
\begin{center}
\includegraphics[width=8.4cm]{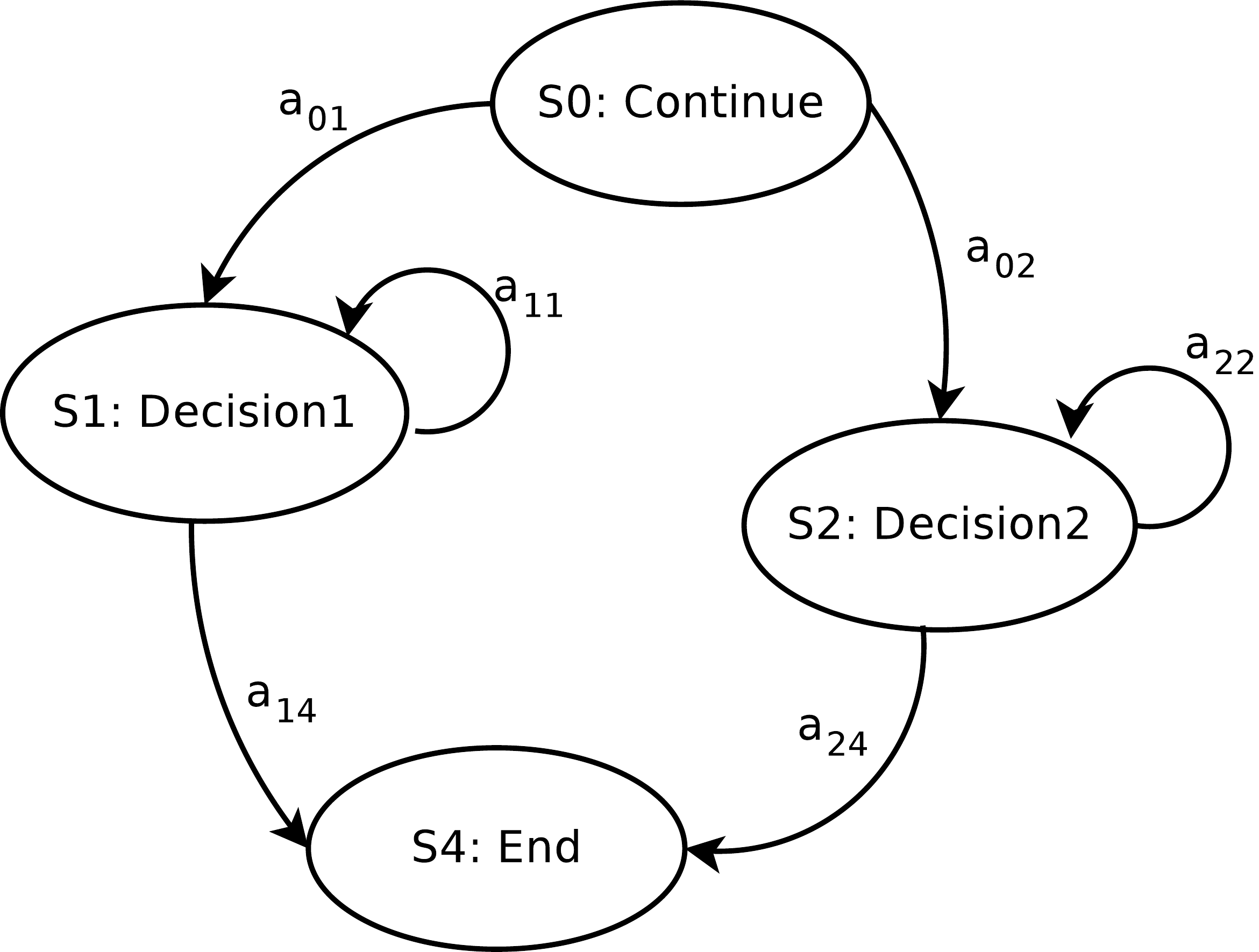}    
\caption{Simplified FSM model for driver with two possible decisions. ``Continue'' state refers to vehicle proceeding with no change.  S1 and S2 could correspond to driver decisions such as stop, turn right. Exemplary transition probabilities are denoted.}  
\label{simple_fsm}                                 
\end{center}                                 
\end{figure}

\subsubsection{Diminishing Likelihood Probabilities}
The first limitation can be understood through a simple example. Consider the hypothetical DSS FSM shown in Figure~\ref{simple_fsm}. In this example, the DSS contains four possible metastates. The transition probabilities between metastates is given by the indicated values. These metastates can refer to any vehicular event of interest such as turning, going straight, changing lanes, etc. The metastate transition matrix, $A$, can be expressed in matrix form as:

\begin{eqnarray}
A  = \begin{bmatrix}
0 & a_{01} & a_{02} & 0 \\ 
0 & a_{11} & 0  & a_{14} \\ 
0 & 0 & a_{22} & a_{24}\\ 
0 &0  & 0 & 1
\end{bmatrix}
\nonumber
\end{eqnarray}

Now consider a long observation sequence in which the system enters states S1 or S2  for an extended period of time. In such a case, the transition probabilities (after $n$ observations) becomes:

\begin{eqnarray}
A^{n}  = \begin{bmatrix}
0 & a_{01} & a_{02} & 0 \\ 
0 & a_{11} & 0  & a_{14} \\ 
0 & 0 & a_{22} & a_{24}\\ 
0 &0  & 0 & 1
\end{bmatrix}^{n} = PD^{n}P^{-1}
\nonumber
\end{eqnarray}

Where \textit{P} is a 4x4 matrix whose columns contain the eigenvectors of \textit{A}, and \textit{D} is a diagonal matrix with the corresponding eigenvalues of \textit{A}. For the above exemplary values, the eigenvalues of \textit{A} are $a_{00}, a_{11}, a_{22}, 1$. With these eigenvalues $D^{n}$ becomes: 

\begin{eqnarray}
D = \begin{bmatrix}
a_{00}^n & 0 & 0 & 0 \\ 
0 & a_{11}^{n} & 0  & 0 \\ 
0 & 0 & a_{22}^{n} & 0\\ 
0 &0  & 0 & 1
\end{bmatrix}
\end{eqnarray}

Since $a_{00}, a_{11}, a_{22}$ are all transition probabilities that are less than 1, for a large $n$, $a_{11}^n, a_{22}^n \rightarrow 0$ and the system will eventually transition from states S1 or S2 to state S4 regardless of the observed CSS estimate. Given driving patterns which often consist of driving straight for a long period of time (think of highway driving), the unintended transition out of a metastate may occur for one in the \textit{Straight} metastate. 

\subsubsection{Including External Information}

To understand the second limitation, consider the HSS+HMM metastate estimation stage.  In this stage, continuous observations of a vehicle are compared with number of competing models (HMM metastates) to determine the most likely driver behavior metastate in the DSS that caused the continuous observed vehicle trajectory. Without additional information, the metastates used for estimating driver behavior are the same as those used for highway driving which are the same as those used for non-intersection road driving, as shown in Figure~\ref{taskgraft_general}. As one may readily understand, the driver intention metastates used for an intersection may not be appropriate when driving on a road far away from any intersections. Fundamentally, an ideal driver behavior estimation system should be able to make use of external information in order to describe a full driving sequence.  At any given time, only a limited set of actions, and thus driver behaviors, are valid. For example, when driving on a limited access highway, making a left turn may not be possible, and when driving on a city street, exiting a highway may not be possible. Including higher level information about the surroundings in the decision making process may be able to provide insight on what is possible at a given time.


\subsection{Overcoming these limitations}

The limitations discussed in the previous section can be overcome by intelligently limiting the length of sequences and including information from external sources in the driver behavior estimation process. There are a number of methods one can use.

One possible method to describe such a full driving sequences is to build some sort of \textit{all-encompassing} metastate that contains all possible driver states under all scenarios and let a system similar to Figure~\ref{taskgraft_general} estimate driver behavior. Such a method will likely struggle with the drawbacks discussed with learning based classifiers in which training data needs to be carefully selected and long sequences need to be manually truncated. While such a technique may provide a solution to the first problem, it does not allow us to include external information in the driver behavior estimation process. 

Another method to describe a full driving sequence could be to dynamically modify the metastates that make up the DSS. The system can also be designed to intelligently modify the observation likelihood probabilities based on external information. 

Dynamically modifying the metastates of the DSS requires the ability to add or remove metastates. The aim of adding or removing metastates from the DSS is to define a DSS that better defines the metastates required to represent vehicle events given current conditions. For example, a DSS corresponding to Highway Driving should include metastates Exit Highway and Enter Highway, whereas a DSS corresponding to near-Intersection driving will not require these metastates but instead require metastates corresponding to Left and Right turns. Any time that the DSS changes, we can modify the likelihood probabilities.

There are numerous scenarios in which DSS modification is necessary and the driver behavior estimation system should be capable of dynamically changing the metastates that comprise the DSS.


\section{Dynamically Modifying the DSS}
\label{dynamicdss}

Dynamic DSS modification provides the HSS+HMM system the ability to dynamically change the metastates that comprise the DSS. Recall that possible driver decisions are encapsulated by metastates in the DSS. In order to dynamically modify the DSS, the operations required for such modifications are adding and removing metastates. Adding a metastate to the DSS is known as \textbf{Metastate Grafting}. Conversely, removing a metastate from the DSS is known as \textbf{Metastate Pruning}. The ability to graft or prune a metastate from the DSS provides the HSS+HMM system the ability to overcome the limitations mentioned in the previous section.

At each graft or prune, the FSM that represent DSS metastate interconnections is changed to account for the additional or lesser metastates. Grafting and pruning of a metastate can occur at any given time but for the purposes of this study it is limited to instances when there is a change in external factors. 

As a driver readily understands, different inputs can imply very different actions under different situations. For example, at an intersection, a right turn signal indicator (in addition to other observations), may lead to a Right Turn state estimate. On the other hand, a right turn signal indicator (in addition to other observations) on a highway, may lead to a Right Lane Change estimate, or if one is near an exit, perhaps an Exit Highway estimate. This estimation may further change with differing traffic and environmental conditions, even when given similar observation sequences. This requires that the HSS+HMM framework be expanded to make use of such information such as the modified architecture in Figure~\ref{taskgraft_systemArchitecture}.

The original formulation of the HSS, the property that discrete driver states of the DSS generate a continuous trajectory which can be observed through easy to measure CSS state estimates, is still maintained. As previously described, the Discrete State System consists of metastates which encompass HMMs that represent driver events of interest. The CSS represents the vehicle dynamics that, given the input of the DSS, generates the observations that decisions are based on. Previously, the discrete driver decision was based entirely on the actions or behavior of a human driver. As depicted in figure~\ref{taskgraft_systemArchitecture}, in order to represent a driver's use of external information in the decision making process, new system interconnections to the HSS depict external factors that may play into a drivers decision. Specifically, the three state systems have been added:

\begin{itemize}
\item \textbf{Roadway Type Condition State System (RCS)}: High level information that describes the road. Roadway type states may include: ``Highway,'' ``Intersection,'' ``Non-intersection,'' etc.
\item \textbf{Environmental Condition State System (ECS)}: High level environmental information. Environmental condition states may include: ``Icy Roads,'' ``Mountainous roads,'' ``Wet roads,'' etc.
\item \textbf{Traffic Condition State System (TCS)}: Information about current or future traffic patterns, congestion, etc.
\end{itemize}

Each of these state systems are represented by a FSM such as that in Figure~\ref{simple_fsm}. State transitions within these FSMs are governed largely by external information such as GPS coordinates, traffic information and/or weather information. While there are many other possible sources for external information, we believe that a majority of the information we process can be contained within these systems.

With the extended architecture, observations of the CSS state are still related to the DSS estimate through the HSS+HMM system. The new state system additions reflect that driver behavior metastates of the DSS can be affected by these state systems. Further, changes in the states of any of these external state systems allows DSS metastates to be change dynamically with changing external conditions. With this extended system, the estimated driver behavior is determined by observations from the CSS, and in determining likely driver decision, information is included from the RCS, ECS and TCS. 

\begin{figure}[t]
\begin{center}
\includegraphics[width=8.4cm]{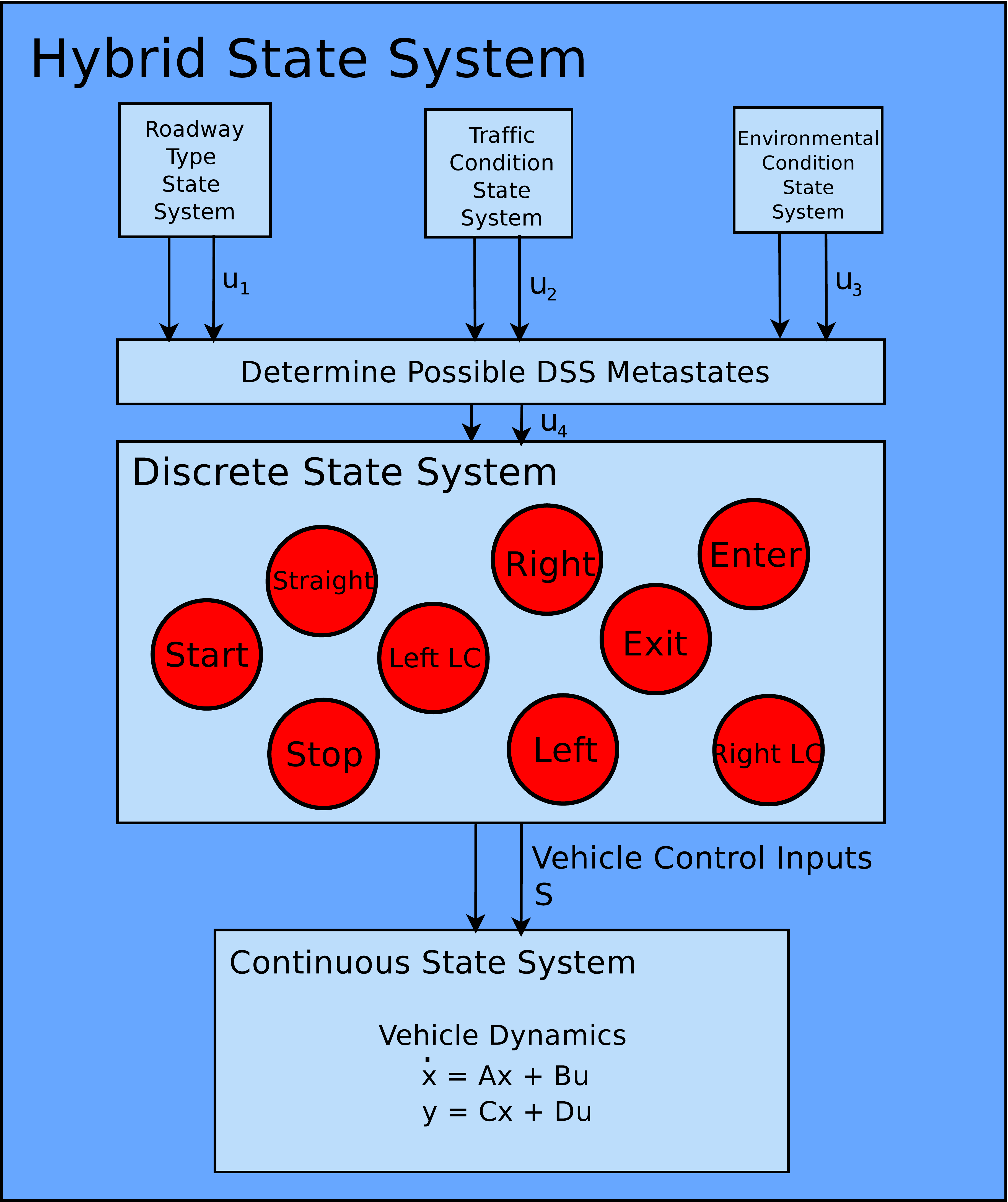}    
\caption{Extended HSS+HMM architecture. Observations measured on a vehicle are the result of driver decisions, road conditions, environmental conditions and traffic conditions.}  
\label{taskgraft_systemArchitecture}                                 
\end{center}                                 
\end{figure}

\subsection{Definitions}
\label{definitions}

With access to high-level information (for example, RCS states can be determined using GPS position), one can dynamically determine a relevant DSS configuration. Given a particular set of states in the RCS, ECS, and TCS a suitable DSS configuration can be used. A DSS configuration is defined to include the following:

\begin{enumerate}
\item Metastate List
\item Metastate FSM (i.e, transition probablities and metastate interconnections)
\item External Information FSM
\end{enumerate}

With this definition, for changes in external conditions, the DSS dynamically changes to model a driver's decision making process. The metastate list represents possible driver behavior events of interest. Consider the DSS, $\textbf{ S}$, which contains metastates $S_{1}, S_{2}, S_{3},..., S_{n}$. For an instance of the DSS $\textbf{S}$ and metastates $S_{i}$, we have the following definitions:

\begin{eqnarray}
\nonumber \textbf{S} \ni \{S_{1}, S_{2}, S_{3}, ... , S_{n}\} \mbox{     }\\
\nonumber |\textbf{S}| = |\{S_{1}, S_{2}, S_{3}, ..., S_{n}\}| = n 
\end{eqnarray}

Where $|S|$ is defined size of $\textbf{S}$ (number of metastates in $\textbf{S}$) . 

The task of modifying DSS metastates is referred to as \textbf{Metastate Modification}. Situations in which a metastate is added to the DSS is referred to as a \textbf{Metastate Grafting}. Situations where a metastate is removed from the DSS is referred to as \textbf{Metastate Pruning}. 

\begin{figure}[h]
\begin{center}
\includegraphics[width=8.4cm]{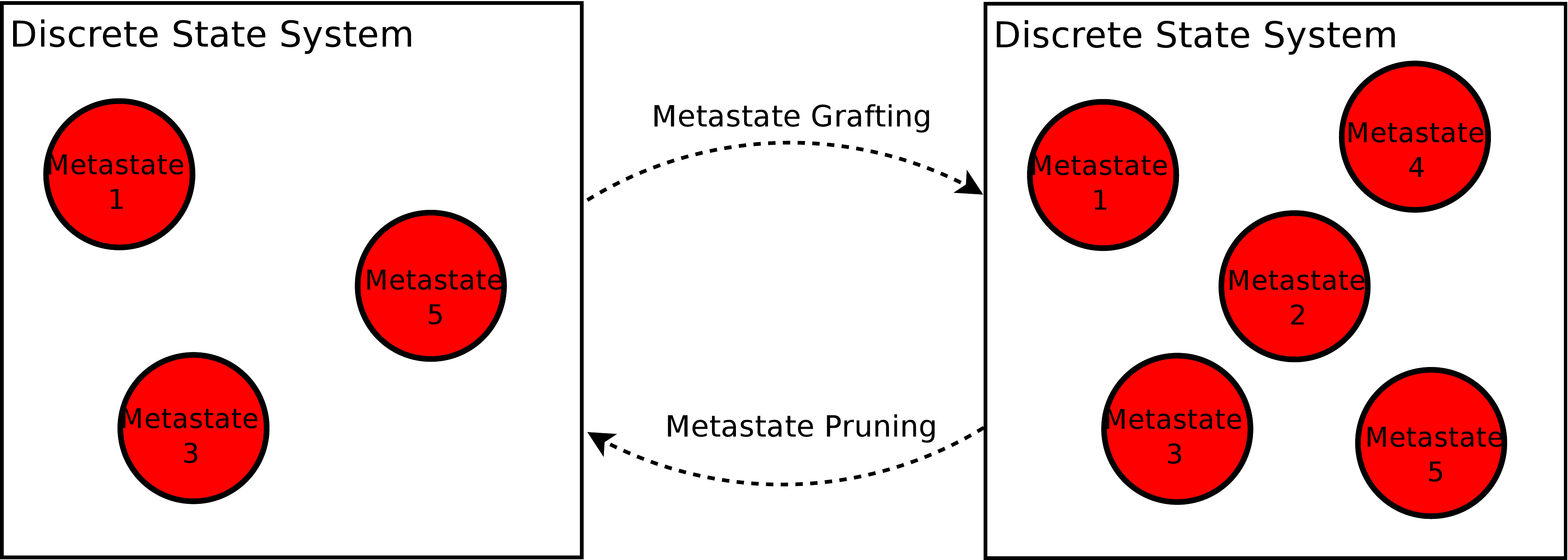}    
\caption{Example Metastate Modification: Left $\rightarrow$ Right shows Metastate Grafting. Right $\rightarrow$ Left shows Metastate Pruning}  
\label{grafting_pruning}                                 
\end{center}                                 
\end{figure}

In this context, we can now mathematically define the two types of Metastate modifications: Metastate Grafting and Metastate Pruning.

\subsubsection{Metastate Grafting}

In a metastate graft, a metastate is added on to the DSS. Let the DSS, \textbf{$S$} at time $t$ be defined as $S(t) \ni \{S_{1}, S_{2}, ... , S_{n}\}$ where $n$ is the number of metastates in the DSS at time $t$. At time $t$, the size of the DSS, $|S| = n$. Now, after $\tau$ time, suppose that the road, environment and/or traffic conditions change such that a new metastate $S_{n+1}$ needs to be grafted into the DSS. Now, $S(t + \tau) \ni \{S_{1}, S_{2}, ... , S_{n}, S_{n+1}\}$.

Formally, the DSS $S$, is said to have been grafted on, if for $S'$ defined as $S \rightarrow_{S_{n+1}}^{graft} S'$, $|S'| = |S| + 1$. Also, since a state has been grafted onto $S$, $S' \supset S$. 

Additionally, if it is necessary to graft more than one state simultaneously, that is the same as serially grafting each individual state. For example:

\begin{eqnarray}\nonumber S \rightarrow_{S_{j}, S_{k}}^{graft} S' \equiv S \rightarrow_{S_{j}}^{graft} \rightarrow_{S_{k}}^{graft} S'
\end{eqnarray}

Consider a simple case where a vehicle is driving on a straight one lane road. In this case, the DSS may contain only one metastate - ``Continue.'' Now, if based on a change in the Road Condition state such as a change from a one lane road to a two lane road, two new metastates will be grafted onto the DSS - ``Left Lane Change'' and ``Right Lane Change.'' Similarly, if the road condition changes to an Intersection, two additional metastates - ``Left Turn'' and ``Right Turn'' can be grafted. 

\subsubsection{Metastate Pruning}

In a metastate prune, a metastate is removed from the DSS based on changes in Road, Environment and/or Traffic conditions. Let the DSS, \textbf{$S$} at time $t$ be defined as $S(t) \ni \{S_{1}, S_{2}, ... , S_{n}\}$ where $n$ is the number of metastates in the DSS at time $t$. At time $t$, the size of the DSS, $|S| = n$. Now, after $\tau$ time, suppose that the road, environment and/or traffic conditions change such that metastate $S_{n}$ needs to be pruned from the DSS. Now, $S(t + \tau) \ni \{S_{1}, S_{2}, ... , S_{n-1}\}$.

Formally, the DSS $S$, is said to have been pruned, if for $S''$ defined as $S \rightarrow_{S_{n}}^{prune} S''$, $|S''| = |S| -1 $. Also, since a state has been pruned from $S$, $S'' \subset S$.

Additionally, if it is necessary to prune more than one state simultaneously, that is the same as serially pruning each individual state. For example:

\begin{eqnarray}\nonumber S \rightarrow_{S_{j}, S_{k}}^{prune} S'' \equiv S \rightarrow_{S_{j}}^{prune} \rightarrow_{S_{k}}^{prune} S''
\end{eqnarray}

Consider a simple case where a vehicle is approaching an intersection. At this stage, the DSS metastates may include ``Stop,'' ``Continue,'' ``Left Turn,'' and ``Right Turn.'' After passing the intersection, the vehicle road condition changes to a single lane road and metastates ``Left Turn'' and ``Right Turn'' are pruned from the DSS.

\subsection{Procedure for Grafting and Pruning}

Based on the current state or change of state of the Roadway Type System (RCS), Environment Condition System (ECS) and Traffic Condition System (TCS), a procedure for grafting and pruning the metastates needs to be defined. For the following examples, we have assumed that the ECS and TCS states are constant. 

Let the state of the RCS at time $t$ be $RC(t)$. Now, suppose that at time $t'$ external information (such as change in GPS position) is received such that the state of the RCS becomes $RC(t')$. Additionally, suppose that the DSS at time $t$ is $S(t) = \{S_{1}, S_{2}, ... , S_{n}\}$. 

If the change in the RCS requires that additional state or states to be added to $S(t)$ to determine $S(t')$, this is represented as follows:

\begin{eqnarray}  S(t) \rightarrow_{metastates}^{graft} S(t') \nonumber \end{eqnarray}

A graft effectively increases the number of possible vehicle events (represented by metastates). Similarly, if the changes in RCS requires the removal of a state or states from $S(t)$ to determine $(S(t')$, this is represented as:

\begin{eqnarray}  S(t) \rightarrow_{metastates}^{prune} S(t') \nonumber \end{eqnarray}

In this case, the prune effectively decreases the number of possible vehicle events. There may be many cases in which $S(t)$ needs more than one graft and/or prune. Suppose that it is necessary to graft $S_{n+1}$ and prune $S_{n}$, this is done through a graft followed by a prune on $S$. 

\begin{eqnarray}  S(t) \rightarrow_{S{n+1}}^{graft} S'(t') \rightarrow_{S_{n}}^{prune} S(t') \nonumber \end{eqnarray}

Where, $S'(t') \ni \{S_{1}, S_{2}, ... , S_{n-1}, S_{n}, S_{n+1}\}$ and $S(t') \ni  \{S_{1}, S_{2}, ... , S_{n-1}, S_{n+1}\}$. With these definitions, given suitable metastates, through the process of grafting and/or pruning various metastates, one can define a DSS that contains any combination of metastates.

Using Figure~\ref{grafting_pruning} as an example. Suppose that the DSS on the left is called $S_{left}$ and the DSS on the right is called $S_{right}$. If the current DSS is $S_{left}$ and changes to external conditions require modifying the DSS to  $S_{right}$, the following operations will need to occur:
\begin{eqnarray} S_{left} \rightarrow_{metastate2}^{graft} \rightarrow_{metastate4}^{graft} S_{right} \nonumber \end{eqnarray}

Similarly in the reverse situation,

\begin{eqnarray} S_{right} \rightarrow_{metastate2}^{prune} \rightarrow_{metastate4}^{prune} S_{left} \nonumber \end{eqnarray}

The grafting and pruning operations defined in this section have been applied to long term driving examples.

\section{Application Examples}
\label{example}

Data collected as described in Section~\ref{datacollection} was used to test the extended HSS+HMM system for long term driver behavior estimation. One of the motivations for introducing high-level information contained in the RCS, ECS, and TCS is to allow for driver behavior estimation of multiple consecutive driver events. In this section, we look at the current implementation of the theoretical system described in Section~\ref{dynamicdss}. We then describe the results obtained of implementing this system on three example driving sequences. To view the second and third examples, you will need a browser capable of playing videos from the online video hosting service YouTube. In all the videos, the state and RC estimate correspond to the vehicle the camera is in.

\subsection{Current Status and Implementation}
\label{currimplementation}

A reverse geocoder, GeoNames (http://www.geonames.org/) is used to offline process the GPS coordinates. This allows us to automatically determine the Roadway Type State for intersections, non-intersections and highways (reverse geocoding for highway segments requires significant manual corrections). \textbf{For the purpose of these examples, it is assumed that the ECS and TCS states remain unchanged} since this information is neither available nor used. Further, RCS states are limited to the following:

\begin{itemize}
\item Intersection $RC_{int}$ (corresponding to DSS $S_{Int}$): This state describes a location close to an intersection (an intersection is defined to be the location of two roads intersecting). $S_{Int} \ni \{S_{Straight}, S_{Left Turn}$ $,S_{Right Turn}, S_{Stop}\}$
\item Highway $RC_{highway}$  (corresponding to DSS $S_{highway}$): This state describes a vehicle operating on a highway or motorway. $S_{highway} \ni \{S_{Straight}, S_{Left Change} , S_{Right Change}\}$
\item Non-intersection $RC_{road}$  (corresponding to DSS $S_{road}$): This state describes a vehicle not described by $RC_{Int}$ or $RC_{highway}$. $S_{road} \ni \{S_{Straight}, S_{Left Change} , S_{Right Change}\}$
\end{itemize}

The grafting and pruning operations are also designed only to occur in situations in which there is a change in the external conditions. A change in external conditions may cause a metastate to be grafted or pruned (or both or any combination thereof) to (or from) the DSS. Whenever there is a change in the composition of the DSS, the current metastate is reset to allow the system to go into any of the available metastates. For example, if the system is in a HMM state of the``Straight/Continue'' metastate and the RCS changes, this leads to metastates being grafted and pruned from the current DSS. At this point, regardless of current metastate, the system will step out of the current metastate and redo the procedure highlighted in Section~\ref{hsshmm} to select the most likely new metastate. 

As further examples, consider a couple of special cases which are mentioned for grafting and pruning operations. In the first case, suppose that the system is currently in a metastate that due to external information has been pruned. In such a case, the system will immediately exit its current metastate and go to a default continue metastate (which is present in every possible DSS configuration) after which the procedure outline before is followed. For the second case, suppose that our system is currently in a metastate when a new metastate is grafted. In this case, the system will essentially reset to allow the ongoing observation to be tested against other metastates including the newly grafted metastate(s). 

A limitation of the current implementation of the system is that a change in the composition of the DSS (i.e, a metastate is grafted or pruned causes the history of the system until that point to be reset as the system \textit{jumping} out of the current metastate. In this reset there is useful information that relates to driving patterns or external conditions that may be lost. In order to extend the methodology presented, it will be beneficial to look at statistical information related to the RCS, but is not presented in this study. For example, collecting information on how RCS changes affect metastate selection.

Even with these limitations to the current implementation of the extended HSS+HMM system, application examples show promise for this theoretical framework.

\subsection{Example 1}
\label{example1}

Using the data collected, an example case that contained multiple consecutive vehicle events was generated. The test observation sequence contains approximately 1900 individual observations captured at a rate of 10Hz. This 2.75 minute observation sequence starts with a vehicle approaching an intersection and stopping at a stop light. After the light turn greens, the vehicle proceeds towards the intersection and turns right. After turning, the vehicle continues straight until coming to a stop at another next intersection. Figure~\ref{grafting_example_20110902} shows the estimation results obtained for this observation sequence as a function of time. The top plot shows the estimated state at a given time step. The lower plot shows the Road Condition State at time $t$, $RC(t) \in \{RC_{Int}, RC_{highway}, RC_{road}\}$. The lowest part of the figure shows the DSS at time $t$. 

For $RC(t) = RC_{Int}$, $S(t) = S_{Int} \ni \{S_{Continue}, S_{Left Turn}$ $,S_{Right Turn}, S_{Stop}\}$. For $RC(t) = RC_{road}$, $RC_{road} \ni \{S_{Continue}, S_{Left Change} , S_{Right Change}\}$

\begin{figure}[h]
\begin{center}
\includegraphics[width=9cm]{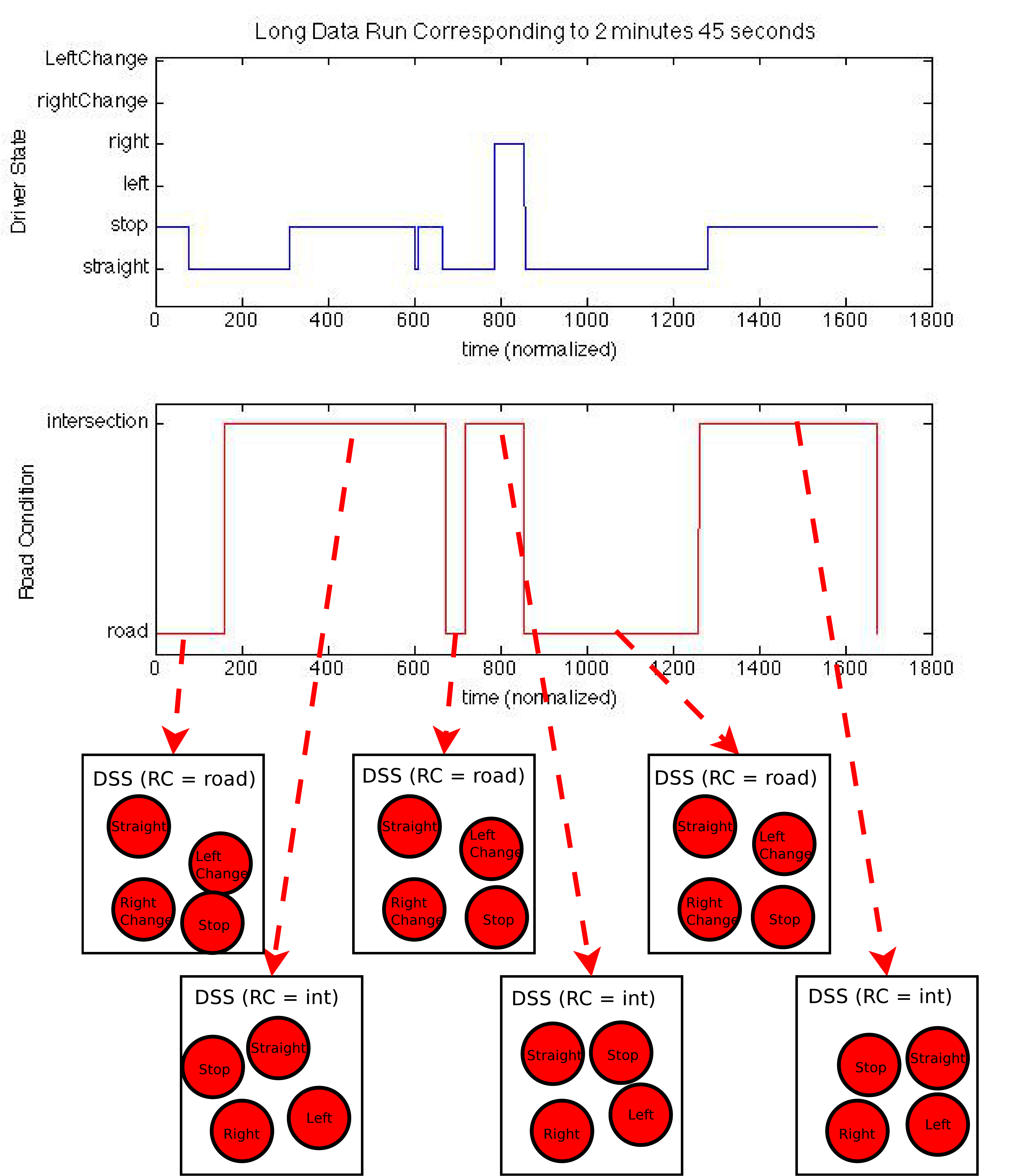}    
\caption{Metastate modification in conjunction used for observation sequence of length 2.75 minutes. For varying road conditions (middle plot) metastates contained in DSS are shown in lowest part of figure.}  
\label{grafting_example_20110902}                                 
\end{center}                                 
\end{figure}

In figure~\ref{grafting_example_20110902}, the grafting/pruning operations when the RCS changes from Road to Int is:

\begin{eqnarray} S_{road} = \{S_{Continue}, S_{Left Change} , S_{Right Change}, S_{Stop}\} \nonumber \\
S_{Int} = S_{road} \rightarrow_{Left Change}^{prune} \rightarrow_{Right Change}^{prune} \rightarrow_{Right}^{graft} \rightarrow_{Left}^{graft} S_{Int}  \nonumber\end{eqnarray}

When the RCS changes from Int to Road the grafting/pruning operations are:

\begin{eqnarray} S_{road} = \{S_{Continue}, S_{Left Turn},S_{Right Turn}, S_{Stop}\} \nonumber \\
S_{road} = S_{int} \rightarrow_{Left Turn}^{prune} \rightarrow_{Right Turn}^{prune} \rightarrow_{Right Change}^{graft} \rightarrow_{Left Change}^{graft} S_{road} \nonumber \end{eqnarray}

\subsection{Example 2}
\label{example2}

Figure~\ref{graftingandgoogleearth} gives the results for another example of metastate modification being used in conjunction with the HSS+HMM to estimate driver behavior. This observation sequence corresponds to approximately 6.5 minutes of city driving. In this time, the driver approaches approximately 10 intersections, which are highlighted in the lower plot of vehicle trajectory using Google Earth.  The upper plot, similar to figure~\ref{grafting_example_20110902}, describes the vehicle state at a given time. The lower plot shows the Road Condition state for the vehicle at a given time. The lower figure shows the vehicle trajectory (white line) of the driver performing the 6.5 minute drive around Columbus, OH. The purple/blue arrows illustrate coordinates where the RCS is Intersection. 

\begin{figure}[t]
\begin{center}
\includegraphics[width=9cm]{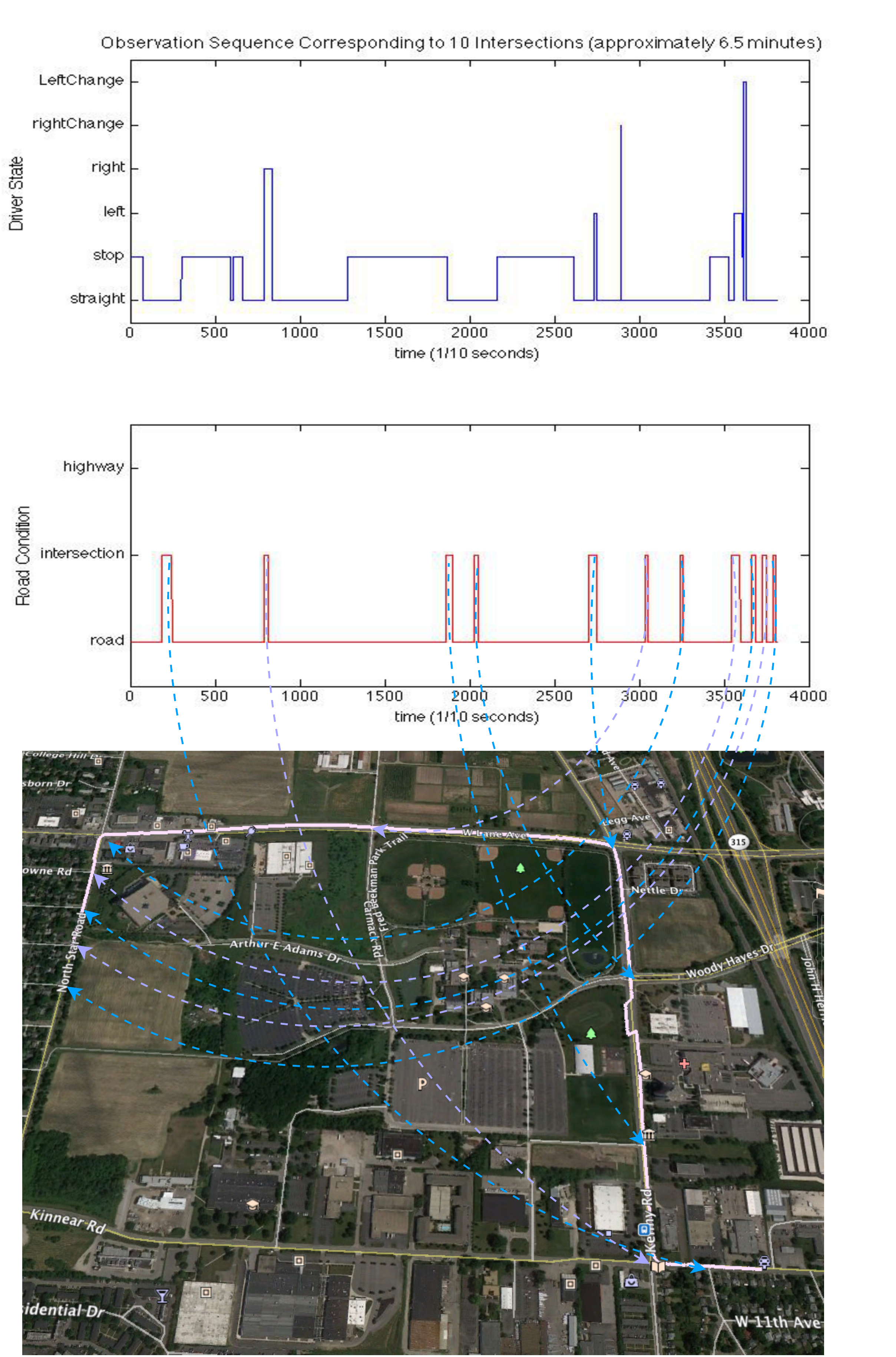}    
\caption{Example of Driver Behavior Estimation using Metastate Modification. Blue and Purple arrows highlight areas where the Road Condition changes}  
\label{graftingandgoogleearth}                                 
\end{center}                                 
\end{figure}

The real-time driver behavior estimation of the sequence of driver behavior depicted in figure~\ref{graftingandgoogleearth} can be viewed online by using the link of Figure~\ref{grafting6min30}. The online video provides the front camera output of the vehicle being driven on the path shown in the lower figure of figure~\ref{graftingandgoogleearth}. The top part of the video gives the current driver state (Event) and road condition (RC) as a 2-tuple. As the vehicle moves, these values are updated using procedures previously outlined. At the beginning of the video (approximately 0:07-0:10 minutes) the driving event is incorrectly shown to be straight, whereas the vehicle is clearly turning right. This difference is due to the RC being Road in which there is no Right Turn metastate. This is one example of the need to have accurate RC, TC, and EC state information. The road condition changes from Road to Int when the distance to an intersection is approximately 100 feet. Starting at 1:56 minutes, a left lane change maneuver is missed. This is largely due to the difficulties of estimating lane changes described in section. Apart from these two anomalies, the driver behavior estimation in the video very accurately estimates driver behavior for changing road conditions.

\begin{figure}[h]
\centering
\includegraphics[width=3cm]{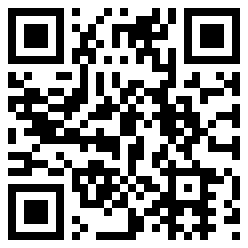}    
\caption{YouTube link~\cite{youtube_6m30s}: Task Grafting with changing road condition information }  
\label{grafting6min30}                                 
\end{figure}

\subsection{Example 3}
\label{example3}

The third example, which you can access by using the link in Figure~\ref{qrcode-1min40secHighwayWithRC}, shows the HSS+HMM behavior estimates for a vehicle operating on the highway over approximately 3.5 minutes. In order to better represent events of interest, this video has been condensed to approximately 1.5 minutes showing only the non-continuous events. In this time, the vehicle perform 4 left lane changes and 3 right lane changes. Some of the segments are prone to a slight delay in lane change recognition largely due to the challenges associated with lane change recognition discussed earlier in this article. The results are displayed in the video screen similar to the second example with the metastate estimate on the right and RC state estimate on the left. 

\begin{figure}[h]
\centering
\includegraphics[width=3cm]{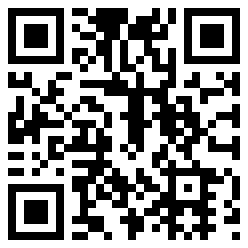}    
\caption{YouTube link~\cite{youtube_1m40s}: Highway Driving }  
\label{qrcode-1min40secHighwayWithRC}                                 
\end{figure}


\section{Conclusions and Future Work}
\label{conclusions}

Cyber Physical Systems have the ability to change human interaction with the world. One particular application area of cyber physical systems, autonomous vehicles or self-driving cars, have the abilty to revolutionize transportation and remedy many of the problems associated with the current state of personal and commercial transporation. In an expected mixed-urban environment, it will be necessary for autonomous vehicles to be able to estimate the driver behavior of a human-driven vehicle. A system for estimating driver behavior has been described in~\cite{gadepally2014framework} but is unable to track the long term behavior of a vehicle. The system proposed in this system extends upon the HSS+HMM system by including external information that shaped driver behavior by dynamically modifying the DSS. Through grafting or pruning operations, the extended HSS+HMM system is capable of emulating driver decisions in changing environments. An implementation of the system has been developed and three examples have been provided. As future work, we can look at improving the computation performance using parallel processing on multi-core or many processors as described in \cite{samsi2010matlab} and \cite{hudak2009computational}.

\section{Acknowledgments}

This material is based upon work supported by the National Science Foundation Grant No. ECCS-0931669. The authors also thank Dr. {\"U}mit {\"O}zg{\"u}ner, Dr. Giorgio Rizzoni, Dr. Keith Redmill, Dr. Albert Reuther, Honda R \& D Americas, Ltd. and the reviewers for their help.

\vfill\break
\bibliography{bibfile}
\bibliographystyle{IEEEtran}

\vspace{10pt}

\begin{IEEEbiography}[{\includegraphics[width=1in,height=1.25in,clip,keepaspectratio]{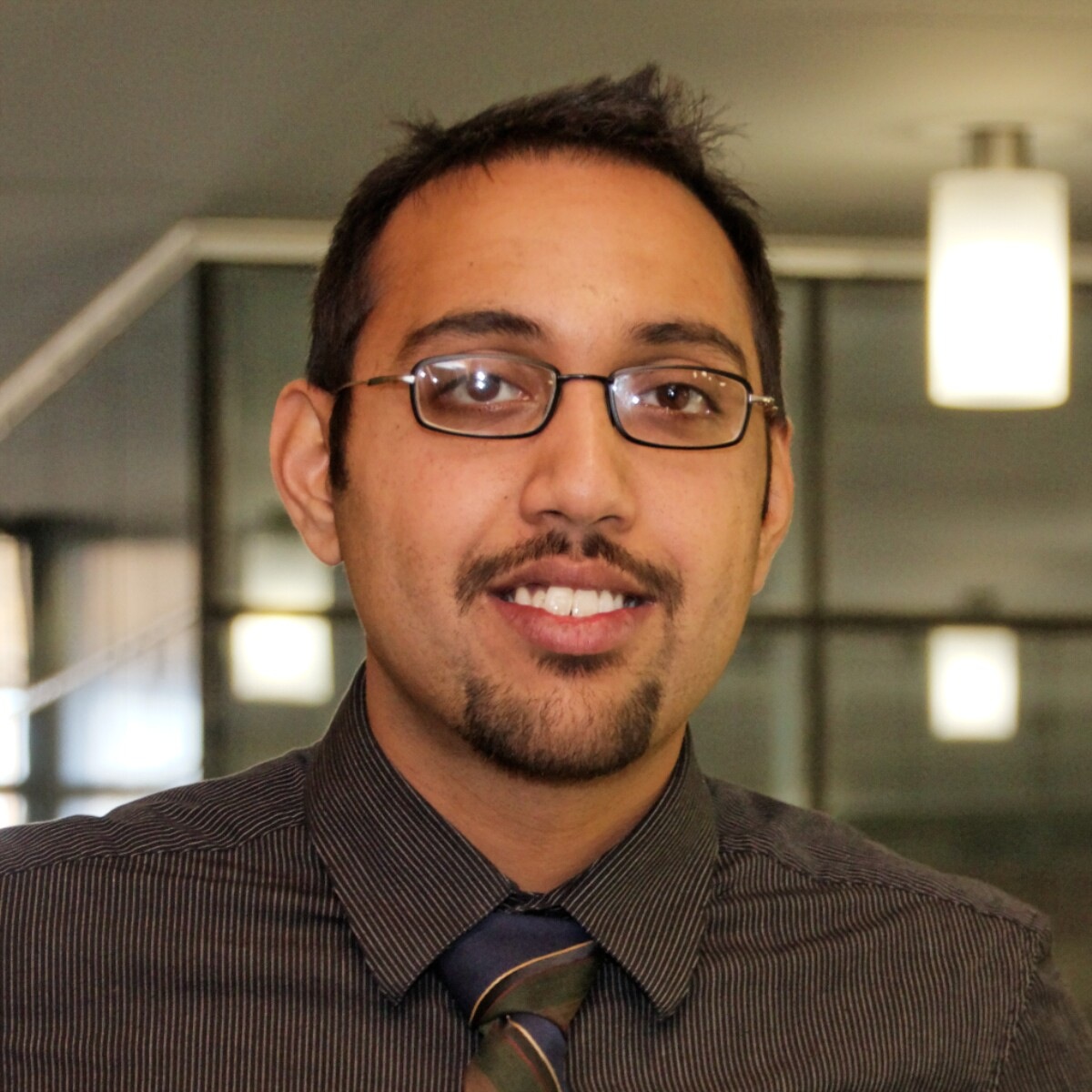}}]{Vijay Gadepally} is a Scientist at MIT Lincoln Laboratory and Computer Science and Artificial Intelligence Laboratory. He earned his bachelors degree in Electrical Engineering from the Indian Institute of Technology, Kanpur and his PhD in Electrical and Computer Engineering from The Ohio State University. He conducts research in autonomous vehicle applications, high performance computing and signal and image processing. 
\end{IEEEbiography}

\vspace{10pt}
\begin{IEEEbiography}[{\includegraphics[width=1in,height=1.25in,clip,keepaspectratio]{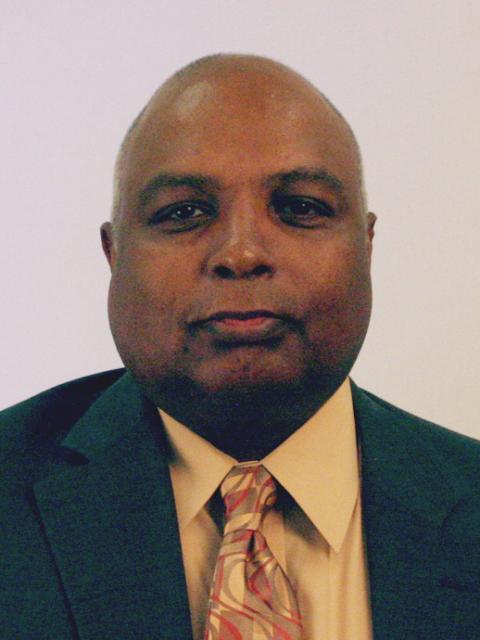}}]{Ashok Krishnamurthy} is the Deputy Director at the Renaissance Computing Institute. He earned his bachelor’s degree in Electrical Engineering from the Indian Institute of Technology in Madras, and his master’s degree and doctorate in electrical engineering at the University of Florida. He is has served as Director of Research and acting co-Director at the Ohio Supercomputer Center and also serves as an Emeritus Professor of Electrical and Computer Engineering department at The Ohio State University. He conducts research in autonomous vehicles, signal/image processing, high performance computing, parallel high-level language applications and computational models of hearing.
\end{IEEEbiography}

\end{document}